\documentclass[letterpaper]{article} 
\usepackage{aaai20}  
\usepackage{times}  
\usepackage{helvet} 
\usepackage{courier}  
\usepackage[hyphens]{url}  
\usepackage{graphicx} 
\usepackage{graphicx}  
\usepackage{amsmath}
\usepackage{amssymb}
\usepackage{multirow}
\usepackage{dsfont}
\usepackage{bm}

\title{Controllable Length Control Neural Encoder-Decoder via \\ Reinforcement Learning}

\author{
Junyi Bian\textsuperscript{\rm 1, \rm 2}, Baojun Lin\textsuperscript{\rm 1}, Ke Zhang\textsuperscript{\rm 1}, Zhaohui Yan\textsuperscript{\rm 1}, Hong Tang\textsuperscript{\rm 1} , Yonghe Zhang\textsuperscript{\rm 2}
\\ \textsuperscript{\rm 1}School of Information Science \& Technology, ShanghaiTech University \\
\textsuperscript{\rm 2}Key Lab of Miscrosatellites, Chinese Academy of Science \\
bianjy@shanghaitech.edu.cn 
}
\begin{document}

\maketitle

\begin{abstract}
    Controlling output length in neural language generation is valuable in many scenarios, especially for the tasks that have length constraints.
    A model with stronger length control capacity can produce sentences with more specific length,
    however, it usually sacrifices semantic accuracy of the generated sentences.
    Here, we denote a concept of Controllable Length Control (CLC) for the trade-off
    between length control capacity and semantic accuracy of the language generation model.
    More specifically, CLC is to alter length control capacity of the model
    so as to generate sentence with corresponding quality.
    This is meaningful in real applications when length control capacity and outputs quality are requested with different priorities,
    or to overcome unstability of length control during model training.
    In this paper, we propose two reinforcement learning (RL) methods
    to adjust the trade-off between length control capacity and semantic accuracy of length control models.
    Results show that our RL methods improve scores across a wide range of target length
    and achieve the goal of CLC.
    Additionally, two models \textit{LenMC} and \textit{LenLInit} modified on previous length-control models are proposed
    to obtain better performance in summarization task while
    still maintain the ability to control length.

\end{abstract}


\section{Introduction}


Neural encoder-decoder was firstly adopted for machine translation \cite{seq2seq_2014}, and fastly diffused to other
domains like image caption \cite{showtell_2015} and text summarization \cite{rush_2015}.
In this paper,  we focus on text summarization
which aims to generate condensed summaries while retains overall points of source articles.
Previous advanced work \cite{rush_2015,beyond_2016} make remarkable progress
and sequence to sequence (seq2seq) framework has become the mainstream in summarization task.
An issue in original neural encoder-decoder is that it cannot generate the sequence with specified length,
i.e., lack of length control (LC) capacity.
Sentences with constrained length are required in many scenarios.
For example, the headlines and news usually have length limit, or
articles and messages in different devices have different length demands.
Generate the sentences with various lengths also improve the diversity of outputs.
However, the study of length control is scarce,
and most research of neural encoder-decoder aim to improve the evaluation score.


To control the output length,
\citeauthor{kikuchi_2016} (\citeyear{kikuchi_2016}) first proposed two learning-based models for neural encoder-decoder
named \textit{LenInit} and \textit{LenEmb}.
We observe that when two models have same or similar structures,
the evaluation score of one model with more precise length control is usually
lower than another with weaker length control.
In other words, worse LC capacity results in better output quality.
For instance, \textit{LenEmb} can generate the sequence with more accurate length but evaluation scores are
lower than \textit{LenInit}.
In most situations when sentence length is in an adequate range, i.e. the length constraint is satisfied,
people prefer to focus on semantic accuracy of the produced sentence,
at this case, \textit{LenInit} seems to be a more appropriate choice than \textit{LenEmb}.
Therefore, it makes sense to research the control of trade-off between LC capacity and sentence quality,
which we called controllable length control (CLC).

To track this trade-off, we set our sight into using Reinforcement Learning (RL) \cite{RL}.
Commonly, RL in neural language generation is used to overcome two issues:
the exposure bias \cite{ranzato_2015} and inconsistency between training objective and evaluation metrics.
Recently, great efforts have been devoted to solve the above two problems
\cite{ranzato_2015,scst_2017,deep_2017}
In addition,
RL can actually bring two benefits in allusion to the LC neural language generation.
Firstly, most datasets provide only one reference summary in each sentence pair,
so we can only learn fixed-length summary for each source document under maximum likelihood (ML) training.
But for RL, we could appoint various lengths as input to sample sentences for training,
consequently, promote the model to become more robust to generate sentences given different desired length.
Secondly, the length information could be easily incorporated into reward design in RL to induce the model to have
different LC capacity, in this way, CLC could be achieved.

Normally, RL for sequence generation is operated on ML-trained models,
however, we find that directly applying RL algorithm on pre-trained models
will dramatically degrade LC capacity.
In this paper, we design two RL methods for LC neural text generation: MTS-RL, and SCD-RL.
By adjusting the rewards in RL according to outputs score and length,
our MTS-RL and SCD-RL can improve the summarization performance as well as control the LC capacity.
Furthermore, we can make some modifications on previous models to improve the score by leveraging the trade-off.
An intuitive approach is that we could add a ``regulator" between length input and decoder to
suppress or enhance the transmission of the length information.
Under the guidance of this idea, two models named \textit{LenLInit} and \textit{LenMC} are proposed.
These two LC models significantly improve the evaluation score at low cost of its ability to control the length in both ML and RL.
The major contributions of our paper are four-fold:

\begin{itemize}
  \item
  To the best of our knowledge, this is the first work applying reinforcement learning on length-control
  neural abstractive summarization, and we present the concept of CLC.
  \item
  Two RL methods are developed to successfully control the LC capacity, and improve the scores significantly.
  Meanwhile, we find that RL for LC text generation alleviate the limitation of inadequacy and unbalance of
  Ground-Truth reference in different lengths.
  \item
  Two models named \textit{LenLInit} and \textit{LenMC}
  are proposed based on previous neural LC models \cite{kikuchi_2016}.
  \item
  Extensive experiments are conducted to verify that proposed models with
  devised RL algorithms cover a wide range of LC ability and smoothly achieve CLC
  on Gigaword summarization Dataset.
\end{itemize}

\section{Related Work}


\subsubsection{Abstractive Text Summarization}

    There are increasing heuristic work based on the encoder-decoder framework \cite{rush_2015,beyond_2016}.
    DRGD designed by \citeauthor{li_2017} (\citeyear{li_2017}) is a seq2seq oriented model equipped with deep recurrent generative decoder.
    \citeauthor{point_2017} (\citeyear{point_2017}) proposed a hybrid pointer-generator network that uses pointer to copy words from articles while produce the words by generator.
    \citeauthor{cao_2018} (\citeyear{cao_2018}) used OpenIE and dependency parser to extract fact descriptions from the source text, then adopted a dual attention model to force the faithfulness of outputs.
    \citeauthor{yang_2019} (\citeyear{yang_2019}) explored a human-like reading strategy for abstract summarization and leveraged it by training model with multi-task learning system.

\subsubsection{Length Control neural Encoder-Decoder}
    \citeauthor{kikuchi_2016} (\citeyear{kikuchi_2016}) first proposed two learning-based neural encoder-decoder models to control sequence length named \textit{LenInit} and \textit{LenEmb}.
    \textit{LenEmb} mixes the inputs of decoder with remaining length embedded
    into each time step, while \textit{LenInit} initializes the memory cell state of LSTM decoder with whole length information.
    Before that, sentence length is controlled by ignoring ``EOS" at certain time or truncating output sentence.
    \citeauthor{fan_2017} (\citeyear{fan_2017}) treated the length of ground truth summaries in different ranges as independent properties and identify it as a discrete mark in an embedding unit.
    \citeauthor{lccnn_2018} (\citeyear{lccnn_2018}) presented a convolutional neural network (CNN) encoder-decoder, the inputs and length information are proceeded by CNN before entering the decoder unit.
    Generally, length control model in neural encoder-decoder can be divided into two types:
    \textbf{Whole Length Infusing (WLI)} model and \textbf{Remaining Length Infusing (RLI)} model.
    WLI model is to inform the decoder with entire length of target sentence and
    RLI model is to tell the remaining length of the sentence in each time step.
    \textit{LenInit} \cite{kikuchi_2016}, Fan \cite{fan_2017} and LCCNN \cite{lccnn_2018} all belong to WLI models,
    while \textit{LenEmb} \cite{kikuchi_2016} is a typical RLI model.
    Ordinarily, RLI models have better length control capacity but lead to poor sentence quality compare with WLI models.
    We follow \citeauthor{kikuchi_2016} (\citeyear{kikuchi_2016}) to define the length of a sentence in character level,
    which is more challenge than \citeauthor{lccnn_2018} (\citeyear{lccnn_2018}) in word level.

\subsubsection{Reinforcement learning in NLG}
    There are several successful attempts to integrate encoder-decoder and RL for neural language generation.
    \citeauthor{ranzato_2015} (\citeyear{ranzato_2015}) applied RL algorithm to directly optimize the non-differential
    evaluation metric, which highly raise score.
    \citeauthor{scst_2017} (\citeyear{scst_2017}) modified RL algorithm by replacing the critic model with inference results to produce rewards,
    this simple modification makes significant improvements in image caption task.
    \citeauthor{seqgan_2017} (\citeyear{seqgan_2017}) rewarded the Monte-Carlo sampled sentences with adversarial trained discriminator.
    \citeauthor{deep_2017} (\citeyear{deep_2017}) employed intra-temporal attention, and combined supervised word prediction with RL to generate more readable summaries.
    \citeauthor{gan_2018} (\citeyear{gan_2018}) designed an adversarial process for abstractive text summarization.
    \citeauthor{fast_2018} (\citeyear{fast_2018}) firstly selected the salient sentences and rewrote the summary,
    in which non-differential computation is connected via policy gradient.
    However, above mentioned work did not involve and explore length control in RL.

\section{Methodology}


\subsection{Problem Definition}
    The dataset $\mathcal{D}$ for text summarization contains pairs of input source sequence $\bm{x} = \{ x_1, x_2, \dots, x_N \}$
    and corresponding ground truth summary $\bm{y}^* = \{ y_1^*, y_2^*, \dots, y_M^* \}$,
    where $N$ and $M$ is the length of the input article and reference, respectively.
    The target of summarization is trying to seek a transform from $\bm{x}$ to $\bm{y}$ using a $\theta$ parameterized policy $p_{\theta}$,
    this can be formalized to maximize the conditional probability in Eq.(\ref{eq1}), where $\bm{y}_{1:t-1}^* = \{y_1^*, y_2^*, \dots, y_{t-1}^*\}$.
    \begin{align} \label{eq1}
        p_{\theta}(\bm{y}^*|\bm{x}) = \prod_{t=1}^M p_{\theta} (y_t^* | \bm{y}_{1:t-1}^*, \bm{x})
    \end{align}



\subsection{Encoder-Decoder Attention Model}
    Encoder-decoder with attention mechanism \cite{bahdanau_2014} is selected as the basic framework in this work.
    RNN encoder sequentially takes each word embedding of input sentence.
    Then the final hidden state of the encoder which contains whole information of source sentence
     is fed into decoder as the initial state.
    We select bi-directional Long Short-Term Memory \cite{lstm} as the encoder to read the source sequence.
    Here we denote $\overrightarrow{\bm{h}}_t^e$ as the hidden state of the BiLSTM encoder in
    forward direction at time step $t$ and $\overleftarrow{\bm{h}}_t^e$ for backward direction.
    $\overrightarrow{\bm{m}}_{t}^e$ and $ \overleftarrow{\bm{m}}_{t}^e$ are the memory cell states of the BiLSTM encoder:
    \begin{equation}
        \label{eq:1}
        \begin{aligned}
            \overrightarrow{\bm{h}}_t^e, \overrightarrow{\bm{m}}_{t}^e  = f_{\text{Bi-LSTM}}^{\text{enc}} ( \overrightarrow{\bm{h}}_{t-1}^e, \overrightarrow{\bm{m}}_{t-1}^e, x_{t-1}) \\
            \overleftarrow{\bm{h}}_t^e, \overleftarrow{\bm{m}}_{t}^e  = f_{\text{Bi-LSTM}}^{\text{enc}} ( \overleftarrow{\bm{h}}_{t+1}^e, \overleftarrow{\bm{m}}_{t+1}^e, x_{t+1})
        \end{aligned}
    \end{equation}
    Outputs of the encoder at time $t$ are concatenated as $\bm{h}_t^e = [\overrightarrow{\bm{h}}_t^e || \overleftarrow{\bm{h}}_t^e]$,
    depicting the vector for attention.
    where $[\cdot || \cdot]$ is denoted as concatenation.

    Decoder unrolls the output summary from initial hidden state by predicting one word each time.
    Neglecting length control, initial state of decoder is set as $\bm{h}_0^d = \overleftarrow{\bm{h}}_{1}^e$ and $\bm{m}_0^d = \overleftarrow{\bm{m}}_{1}^e$,
    and the hidden state $\bm{h}_t^d$ is calculated by:
    \begin{equation}
        \bm{h}_t^d,  \bm{m}_{t}^d = f_{\text{LSTM}}^{\text{dec}}  (\bm{h}_{t-1}^d, \bm{m}_{t-1}^d, x_{t-1})
    \end{equation}
    Context vector $\bm{c}_t$ is used to measure which parts of the source words that decoder pays attention to at time $t$:
    \begin{align}
        e_{t,i} =& \bm{v}^T \tanh (W_{ad} \bm{h}_t^d + W_{ae} \bm{h}_i^e + b) \\
        \bm{c}_t =& \sum_{i=1}^N \alpha_{t,i} \bm{h}_i^e,\ \ \ \ \alpha_{t,i} = \frac{\exp(e_{t,i})}{\sum_{j=1}^N \exp(e_{t,j})}
    \end{align}
    Then we can concatenate $\bm{c}_t$ with hidden state $\bm{h}_t^d$ to
    predict the next word:
    \begin{align}
        p(y_{t}|y_{1:t-1}, \bm{x}) = \text{softmax} \left( W_{p}[\bm{h}_t^d \parallel \bm{c}_t] + b \right)
    \end{align}


\subsection{Length Control Models}
    \begin{figure}[t]
    \centering
    \includegraphics[width=0.9\columnwidth]{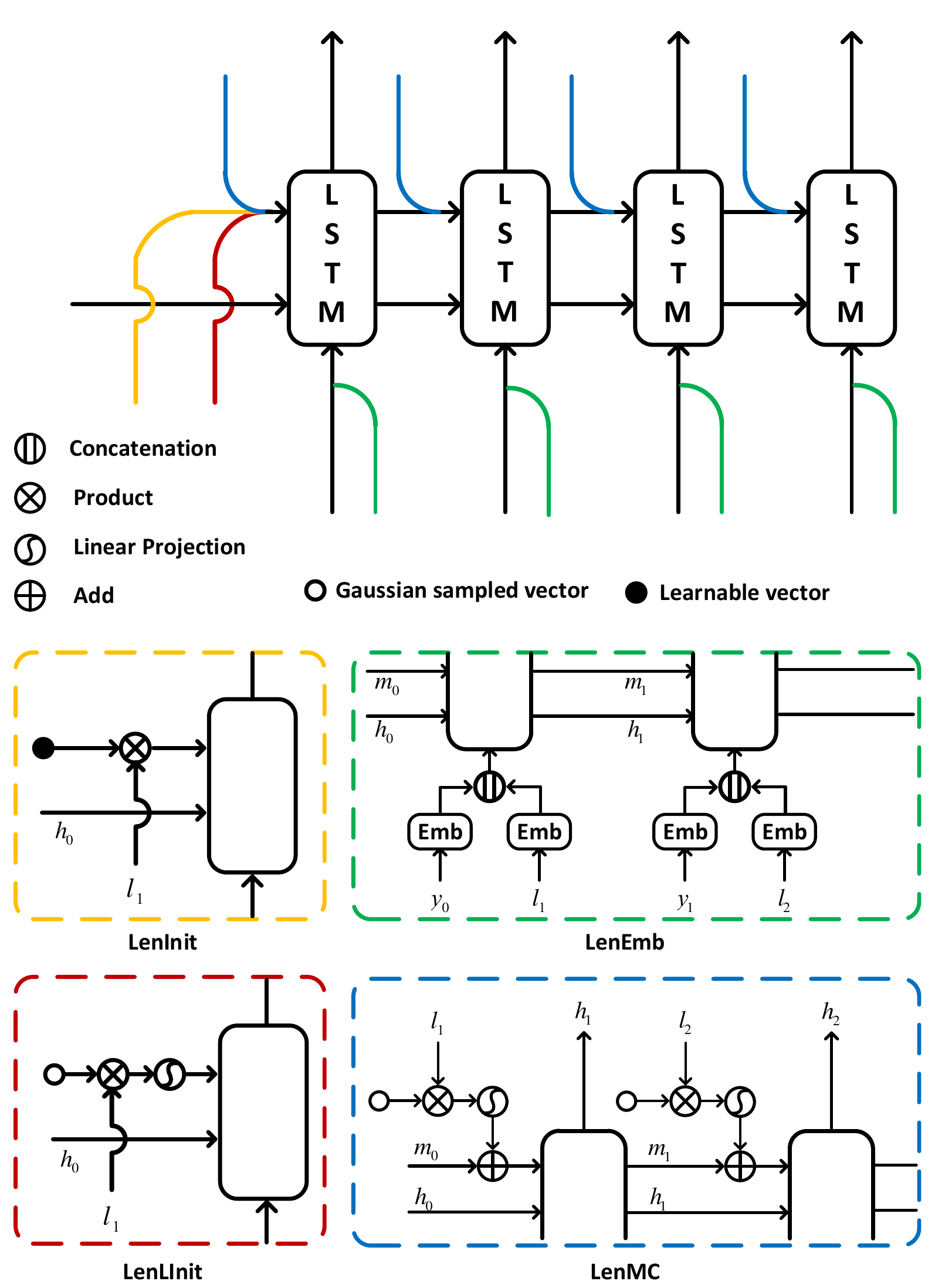} 

    \caption{Decoder structure of four neural length-control models.
    Four colors in above area represent the different modification of original LSTM in four models, respectively.
    Below are details of the corresponding model structures.}
    \label{fig1}
\end{figure}

    To control the length of output, we need to put the desired length information into the decoder,
    hence, the training objective in supervised ML with ``teacher forcing"\cite{teacher_1989} becomes:
    \begin{align}
        L_{ml}(\theta) = -\sum_{t=1}^M \log p_{\theta} (y_t^* | \bm{y}_{1:t-1}^*, \bm{x}, l_{t})
    \end{align}
    Here, $l_{t}$ is denoted as length information the decoder perceives at time $t$.
    As is introduced before, LC models are classified into two groups.
    For the RLI model, remaining length is updated in each time step by $l_{t+1} = l_{t} - \text{len}(y_t^*)$, while $l_1$ is set to $\text{len}(\bm{y}^*)$.
    In WLI model, decoder only aware of the whole length of the sentence, so we set all $l_t$ as $\text{len}(\bm{y}^*)$.

    In this section, We will introduce four models:
    \textit{LenInit}, \textit{LenEmb}, \textit{LenLInit} and \textit{LenMC}.
    The first two models are proposed by \cite{kikuchi_2016}.
    We make modification on them and propose the remaining two.
    \subsubsection{LenInit}
        This WLI model uses memory cell to control the output length by rewriting the initial state $\bm{m}_0^d$ as:
        \begin{align}
            \bm{m}_0^d = \bm{b}_l * l_1
        \end{align}
        $l_1$ is regarded as the entire desired length of the output sentence,
        and $\bm{b}_l \in \mathbb{R}^D$ is a learnable vector.

    \subsubsection{LenLInit}
        This model can be viewed as a variant of \textit{LenInit}.
        In order to produce higher scores by leveraging the LC capacity,
        we simply add a linear transformation $W_l$ of length information,
        the model is thus named Length Linear Initialization (\textit{LenLInit}).
        Unlike the \textit{LenInit},
        $\bm{b}_l$ is replaced by $\bar{\bm{b}}$, a gaussian sampled non-trainable vector,
        and the initial memory cell state of decoder is:
        \begin{align}
            \bm{m}_0^d = W_l(\bar{\bm{b}} * l_1)
        \end{align}

    \subsubsection{LenEmb}
        For this RLI model, embedding matrix $W_{le}\in \mathbb{R}^{D\times L}$
        transforms $l_{t}$ into a vector $\bm{e}_l(l_{t}) \in \mathbb{R}^D$, where $L$ is the possible length types,
        then $\bm{e}_l(l_{t})$ will be concatenated with the word embedding vector $\bm{e}_w(y_{t-1}^*)$ as additional input for LSTM decoder:
        \begin{align}
            \bm{h}_t^d, \bm{m}_t^d = f_{\text{LSTM}}^{\text{dec}}(\bm{h}_{t-1}^d, \bm{m}_{t-1}^d, [\bm{e}_w(y_{t-1}^*) \parallel \bm{e}_l(l_{t})])
        \end{align}

    \subsubsection{LenMC}
        Other than \textit{LenEmb} that length information $\bm{e}_l(l_{t})$ is concatenated as additional input,
        we infuse $l_{t}$ into memory cell at each time step in the same way as \textit{LenLInit}, and name this RLI model as \textit{LenMC}.
        \begin{align}
            \bm{h}_t^d, \bm{m}^t_d = f_{\text{LSTM}}^{\text{dec}}(\bm{h}_{t-1}^d, \bm{m}_{t-1}^d + W(\bar{\bm{b}} * l_{t})  , e_w(y_{t-1}^*))
        \end{align}

    \begin{figure}[h]
    \centering
    \includegraphics[width=0.9\columnwidth]{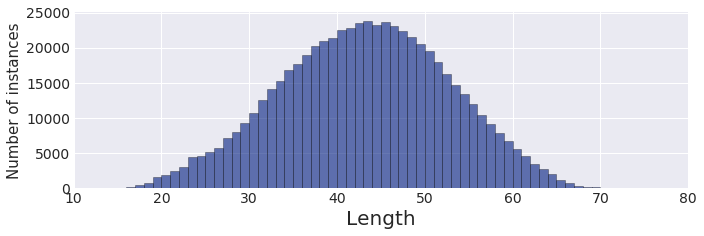} 

    \caption{Summary length distribution in parts of Gigaword}
    \label{fig2}
\end{figure}

\subsection{Length Control Reinforcement Learning}

    Models trained by maximum likelihood estimation with ``teacher forcing" suffer from the problem of ``exposure bias" \cite{ranzato_2015}.
    Moreover, the training process is to minimize the cross-entropy loss, while in test time, results are evaluated with language metrics.
    One way to redeem these conflicts is to learn a policy that directly maximizes the evaluation of metric instead of
    maximum-likelihood loss where RL could be naturally considered.

    From the perspective of RL for sequence generation,
    our LC models can be viewed as an agent, parameters of the network form a policy $p_{\theta}$,
    and making prediction at each step can be treated as action.
    After generating a complete sentence,
    agent receives a reward computed by evaluation metrics.
    During training process, decoder can produce two types of output:
    $\bm{y} = \{ y_1, y_2, \dots \}$ with greedy search,
    and $\bm{y}^s = \{ y_1^s, y_2^s, \dots \}$ in which $y_t^s$ is sampled from the probability distribution
    $p_{\theta}(y_t^s|y_{1:t-1}^s, \bm{x})$ at time $t$.
    We assign a random number $l_t^s$ within an appropriate range as the target summary length for each article
    and feed it into LC model to sample a sentence $\bm{y}^s$,
    then reward $r(\bm{y}^s)$ is evaluated between ground truth summary $\bm{y}^*$ and sampled sentence $\bm{y}^s$.
    We apply the self-critical sequence training (SCST) \cite{scst_2017} as our RL backbone,
    and the training objective of SCST becomes:
    \begin{align}
        L_{rl}(\theta) = (r(\bm{y}) - r(\bm{y}^s)) \sum_{t=1} \log p_{\theta} (y_t^s|\bm{y}_{1:t-1}^s, \bm{x}, l_t^s)
    \end{align}
    This reveals that the goal of policy gradient RL in sequence generation is equivalent to
    increase the probability of generating high-score sentences.

    We encounter two additional problems about LC summarization,
    first is that LC models are designed to generate summaries in different lengths,
    but existing datasets only provide one or a few ground-truth references for each article,
    worse still, the number of reference with different length are terribly unbalanced (see Figure \ref{fig2}).
    In consequence, models trained under ML by this dataset tend to have better performance only in particular lengths.
    By sampling sequences with randomly assigned length $l_1^s$ in reinforcement training,
    uniform-distributed length sentences are served as additional summaries to be judged by
    RL system, as a result, alleviate the above-mentioned issue.


    The second problem is that directly applying SCST for LC models will seriously diminish the LC capacity,
    because some of the sampled sentences have deviation in length,
    enlarging the generation probability of these sentences will corrupt the LC capacity
    that in turn would further force the model to generate more length-deviation sentences,
    and therefore reinforcing a vicious cycle to lead LC capacity crash.
    To save the model from length control collapse in RL,
    an intuitive idea is to adjust the reward incorporating with outputs length,
    especially for those sentences with high scores and mismatched length.
    In consequence, we propose two training approaches for length control RL:
    Manually Threshold Select (MTS) and Self-Critical Dropout (SCD).
    Additionally, both training algorithms can regulate the model by tuning a hyper-parameter that
    has better LC capacity but lower sentence quality and vice versa,
    i.e, accomplishing the CLC.

    \subsubsection{Manually Threshold Select}
        As an initial point, semantic accuracy is still the most critical indicator needed to be guaranteed.
        For a sentence has low score, its generation probability would be reduced during the training
        even with expected length.
        Considering sentences with high scores,
        for those who have expected length, reward should be naturally retained,
        thus, we only need to deal with remaining sentences with unqualified length.

        Suppose the desired length for sampling sentence is $l_1^s$, and the length of the output sequence is ${len}(\mathbf{y}^s)$.
        The length prediction error $d_e$ is the difference of two lengths: $abs(l_1^s - len(\bm{y}^s))$.
        We manually choose an error threshold $d_{th}$
        to eliminate the reward of sentence when $d_e$ exceeds $d_{th}$:
        \begin{equation}
            r(\bm{y}^s) =\left \{ \begin{array}{lll} r(\bm{y}) & r(\bm{y}^s) > r(\bm{y})\ \text{and}\ d_e > d_{th} \\ r(\bm{y}^s)  & \text{otherwise} \end{array} \right.
        \end{equation}

        The LC capacity can be adjusted by setting different $d_{th}$, larger $d_{th}$ would yield better evaluation score
        while smaller $d_{th}$ get better length control.

    \subsubsection{Self-Critical Dropout}
        Two drawbacks occur in MTS-RL.
        Firstly, sentences exceeding the limit are completely ignored even though they reach high evaluation scores while $d_e$
        is slightly larger than $d_{th}$.
        Secondly, $d_{th}$ can only take discrete values,
        and this makes it hard to control those models that have precise length control such as \textit{LenMC}.
        Inspired by SCST \cite{scst_2017} to approximate the baseline from the current training model,
        we propose Self-Critical Dropout RL approach.
        In each iteration, a batch of sampled outputs
        $\mathcal{B} = \{\bm{y}^{s, 1}, \bm{y}^{s, 2}, ..., \bm{y}^{s, |\mathcal{B}|}\}$ is obtained,
        where $\bm{y}^{s, i}$ is the $i^{th}$ sampled sentence with desired length $l_1^{s, i}$.
        The mean of $d_e$ is approximated by:
        \begin{equation}
            \bar{d}_e \approx \frac{1}{|\mathcal{B}|} \sum_{i=1}^{|\mathcal{B}|} \text{abs} \left( \text{len}(\bm{y}^{s,i}) - l_1^{s, i} \right)
        \end{equation}

        We take $\bar{d}_e$ as the threshold,
        unlike the previous method that restrains the rewards of all sentences with $d_e$ larger than $d_{th}$,
        we keep their rewards by a probability of $p_{\text{select}}$.
        At the same time, rewards should be more likely to be reserved when $d_e$ get closer to $\bar{d}_e$:
        \begin{align}
            p_{\text{select}} = \exp (-\lambda (d_e - \bar{d}_e))
        \end{align}
        $\lambda$ reflects the degree of length constraint towards output sequence,
        therefore controls the LC capacity.
        Larger $\lambda$ could force the model to generate sentences that have more accurate lengths,
        while smaller $\lambda$ have weaker control of length so could improve the performance.






\begin{table*}[ht]
    \centering
    \caption{Example summaries of four LC models.
    (Note that ``gunners" is a nickname of arsenal)} \smallskip
    \begin{tabular}{l|l|l}
        \hline \hline
        Source article & \multicolumn{2}{p{12cm}}{\raggedright arsenal chairman peter hill-wood revealed thursday that he fears french striker thierry henry will leave highbury at the end of the season .} \\ \hline \hline
        Reference summary & \multicolumn{2}{p{12cm}}{arsenal boss fears losing henry} \\ \hline \hline
        \multicolumn{3}{c}{} \\ \hline
    model      & \multicolumn{2}{l}{desired length and sampled summaries (true length)} \\ \hline
               & 25 & arsenal chief quits to leave (24) \\ \cline{2-2}
    LenLInit   & 45 & gunners chief fears french striker will leave the end (45) \\ \cline{2-2}
               & 65 & gunners chief says he will leave as he fears french striker will leave (58) \\ \hline \hline
               & 25 & arsenal fears henry henry (22) \\ \cline{2-2}
    LenInit    & 45 & arsenal fears french striker henry will leave arsenal (46) \\ \cline{2-2}
               & 65 & arsenal fears french striker henry will leave arsenal says arsenal chairman (65) \\ \hline \hline
               & 25 & arsenal fear henry will leave (25) \\ \cline{2-2}
    LenMC      & 45 & arsenal 's arsenal worried about henry 's return home (45) \\ \cline{2-2}
               & 65 & arsenal 's arsenal worried about french striker henry will leave wednesday (64) \\ \hline \hline
               & 25 & arsenal 's henry to quit again (25) \\ \cline{2-2}
    LenEmb     & 45 & arsenal chairman fears henry 's fate of henry 's boots (45) \\ \cline{2-2}
               & 65 & arsenal chairman fears french striker henry says he 's will leave retirement (65) \\ \hline
    \end{tabular}
    \label{table1}
\end{table*}

\section{Experiments}
    The experiments are divided into two parts.
    We make basic experiments in ML to observe the gap of accuracy between LC models and other summarization baselines.
    Besides, trained models will be served as the initial state of RL.
    Then further comparison on LC models under different RL methods is conducted,
    we pay more attention to this part and perform extensive experiments to demonstrate the effectiveness of controllable length control by designed RL.

\subsection{Experiment Setting}

    \subsubsection{Gigaword Dataset}

        Gigaword dataset is selected for our experiments. The corpus pairs including
        the collected news and corresponding headlines \cite{giga_2012}.
        We use the standard train/valid/test data splits followed by \cite{rush_2015},
        which are pruned to improve data quality.
        The whole processed dataset contains nearly 3.8 million sentences for training, along with one summary each.
        In the experiment of ML,
        to compare with other summarization models in a unified standard,
        we conduct the experiment on the entire dataset.
        Results are reported by standard Gigaword testset which contains 1951 instances and
        we name it ``test-1951".
        For the experiments on RL, we shrink the size of training set by sampling 600K pairs of it,
        validation/test set is rebuilt imitating \citeauthor{infused_2018} (\citeyear{infused_2018}),
        two non-overlapped sets are sampled from a standard validation set called: ``valid-10K" and ``test-4k"
        for model selection and result evaluation, respectively.

        Notice that the scores on ``test-4k" are much higher than those on ``test-1951",
        this is because in standard test set, words in summary sentences do not frequently occur in source texts
        which brings difficulty for word prediction during decoding.
        We build the dictionary containing 50000 words with the highest frequency and
        the other words are replaced by ``unk" tag.

%

\begin{table}[h]
  \centering
  \caption{Results of ML training on standard ``test-1951"} \smallskip
    \begin{tabular}{|l|ccc|c|}
    \hline
    model name &     R-1     &     R-2    &     R-L     &  svar \\ \hline \hline
    \multicolumn{5}{|c|}{Summarization models} \\ \hline
    ABS                &       29.55     &      11.32     &     26.42    & -  \\
    ABS+               &       29.76     &      11.88     &     26.96    & -  \\
    Luong-NMT          &       33.10     &      14.45     &     30.71    & -  \\
    RAS-LSTM           &       32.55     &      14.70     &     30.03    & -  \\
     RAS-ELman         &       33.78     &      15.97     &     31.15    & -  \\ \hline
    seq2seq (our impl.)      &       32.24     &      14.92     &     30.21    &  14.23  \\ \hline \hline
    \multicolumn{5}{|c|}{Length-control models} \\ \hline
     LenLInit (our)     &       \textbf{30.47}     &      \textbf{13.35}     &     \textbf{28.40}    &  2.15  \\
     LenInit           &       29.97     &      13.03     &     28.07      &  2.11  \\
     LenMC (our)       &       29.45     &      12.65     &     27.41      &  0.87  \\
     LenEmb            &       28.83     &      11.89     &     26.92      &  \textbf{0.85}  \\ \hline
    \end{tabular}
    \label{table2}
\end{table}

    \subsubsection{Evaluation Metric}
        Following other summarization work, we evaluate the quality of generated sentence by F-1 scores of ROUGE-1(R-1), ROUGE-2(R-2), ROUGE-L(R-L) \cite{rouge_2004}.

        To measure the LC capacity,
        \citeauthor{lccnn_2018} (\citeyear{lccnn_2018}) use variance of summary lengths $\text{len}(\bm{y})$ against target length $l_1$,
        In this paper, we use the square root of variance (svar) :
        \begin{align}
            svar = \sqrt{\sum_{i=1}^{|\mathcal{D}|} \frac{1}{|\mathcal{D}|} (\text{len}(\bm{y}^i) - l_1^i)^2}
        \end{align}

        \begin{table*}[ht]
    \centering
    \caption{Performance of length control RL in ``test-4k" (ML results also included for comparison).
    Obviously highest scores (0.4 larger than the second best) are in bolded font,
    the scores in italic font are significantly worse score (2 lower than best socre).
    } \smallskip
    \begin{tabular}{ll|ccc|ccc|ccc|c}
        \hline
                      &  & \multicolumn{3}{c|}{25} & \multicolumn{3}{c|}{45} & \multicolumn{3}{c|}{65} &  \\ \cline{3-11}
        model         & parameter & R-1 & R-2 & R-L         & R-1 & R-2 & R-L         & R-1 & R-2 & R-L         & svar($\pm$std) \\ \hline \hline
        \multicolumn{12}{c}{ML} \\ \hline
        LenLInit         & & \textbf{39.03} & \textbf{17.68} & \textbf{37.46}   & 42.04 & 20.47 & 39.87   & \textbf{39.40} & \textbf{18.71} & \textbf{36.96}   & 3.96 \\
        LenInit          & & 37.36 & 16.76 & 35.92   & 42.11 & 20.55 & 39.83   & 38.67 & 18.23 & 36.33   & 2.98 \\
        LenMC            & & 37.10 & 16.68 & 35.72   & 41.38 & 19.98 & 38.99   & 37.93 & 17.87 & 35.51   & 1.05 \\
        LenEmb           & & \textit{34.77} & \textit{14.85} & \textit{33.41}   & \textit{40.00} & \textit{18.43} & \textit{37.74}   & \textit{36.96} & \textit{16.89} & \textit{34.49}   & \textbf{0.97} \\ \hline \hline
        \multicolumn{12}{c}{SCST } \\ \hline
        LenLInit       &    & \textbf{42.90} & \textbf{20.10} & \textbf{40.78}  &  \textbf{43.48} & \textbf{20.83} & \textbf{40.99}  &  \textbf{42.61} & \textbf{20.37} & \textbf{40.06} & \textit{11.0$\pm$2.57} \\
        LenInit        &    & 39.55 & \textit{17.45} & \textit{37.85}  &  42.75 & 20.20 & 40.35  &  40.79 & 19.01 & 38.37 & \textit{8.90$\pm$2.12} \\
        LenMC          &    & 40.38 & 18.14 & 38.52  &  42.14 & 19.98 & 39.48  &  \textit{38.36} & \textit{17.75} & \textit{35.65} & 2.46$\pm$0.47\\
        LenEmb         &    & \textit{37.77} & \textit{15.42} & \textit{35.88}  &  \textit{40.40} & \textit{18.24} & \textit{37.75}  &  \textit{37.48} & \textit{16.87} & \textit{34.77} & \textbf{1.59$\pm$0.12} \\ \hline \hline
        \multicolumn{12}{c}{MTS-RL } \\ \hline
        LenLInit & $d_{th}=16$ & \textbf{42.64} & \textbf{20.13} & \textbf{40.50}  &  \textbf{43.12} & 20.80 & \textbf{40.62}  &  \textbf{41.44} & \textbf{19.81} & \textbf{38.91} & \textit{8.54$\pm$0.78} \\
        LenLInit & $d_{th}=8$  & 41.43 & 19.01 & 39.46  &  42.63 & 20.55 & 40.23  &  39.81 & 19.03 & 37.43 & 5.14$\pm$0.60 \\
        LenLInit & $d_{th}=4$  & 40.66 & 18.43 & 38.85  &  42.46 & 20.45 & 40.02  &  39.13 & 18.61 & 36.70 & \textbf{3.87$\pm$0.10} \\ \hline
        LenInit & $d_{th}=16$  & \textbf{40.22} & \textbf{17.88} & \textbf{38.42}  &  \textbf{42.77} & 20.36 & \textbf{40.31}  &  \textbf{40.32} & \textbf{18.83} & \textbf{37.69} & \textit{6.17$\pm$0.46} \\
        LenInit & $d_{th}=8$   & 39.52 & 17.75 & 37.79  &  42.42 & 20.19 & 39.95  &  39.16 & 18.28 & 36.57 & 3.50$\pm$0.56 \\
        LenInit & $d_{th}=4$   & 38.62 & 17.31 & 36.98  &  42.26 & 20.29 & 39.82  &  38.52 & 18.00 & 36.04 & \textbf{2.79$\pm$0.13} \\ \hline
        LenMC   & $d_{th}=1 $  & 38.56 & 16.53 & 36.89  &  41.33 & 19.67 & 39.04  &  37.83 & 17.66 & 35.39 & 1.01$\pm$0.07 \\
        LenMC   & $d_{th}=0 $  & 38.60 & \textbf{16.98} & 36.98  &  \textbf{41.89} & 20.08 &  39.37 &  38.18 & 17.93 & 35.68 & 0.89$\pm$0.02 \\ \hline \hline
        \multicolumn{12}{c}{SCD-RL } \\ \hline
        LenLInit & $\lambda=0.1$  & \textbf{41.22} & \textbf{18.95} & \textbf{39.31}  &  \textbf{42.77} & 20.65 & \textbf{40.34}  &  \textbf{40.14} & \textbf{19.22} & \textbf{37.76} & 6.05$\pm$0.45 \\
        LenLInit & $\lambda=0.8$  & 40.12 & 18.15 & 38.41  &  42.25 & 20.46 & 39.93  &  39.12 & 18.59 & 36.67 & \textbf{3.84$\pm$0.05} \\ \hline
        LenInit  & $\lambda=0.1$  & \textbf{40.45} & \textbf{17.88} & \textbf{38.48}  &  \textbf{42.88} & 20.08 & \textbf{40.30}  &  \textbf{40.26} & \textbf{18.47} & \textbf{37.38} & 4.52$\pm$0.14 \\
        LenInit  & $\lambda=0.8$  & 38.39 & 17.07 & 36.75  &  42.14 & 20.21 & 39.75  &  38.45 & 17.98 & 35.88 & \textbf{2.64$\pm$0.06} \\ \hline
        LenMC    & $\lambda=0.1$  & \textbf{39.86} & \textbf{17.69} & \textbf{38.13}  &  \textbf{42.27} & 20.19 & \textbf{39.78}  &  \textbf{38.33} & 18.02 & \textbf{35.78} & 1.46$\pm$0.04 \\
        LenMC    & $\lambda=0.4$  & 38.87 & 17.28 & 37.28  &  41.64 & 19.95 & 39.20  &  37.75 & 17.83 & 35.41 & \textbf{1.15$\pm$0.02} \\ \hline

    \end{tabular}

    \label{table3}
\end{table*}

    \subsubsection{Implementation details}
        Dimensions of hidden state for our BiLSTM encoder and one-layer LSTM decoder are both fixed to 512.
        The size of vector $\bm{b}_l$ and $\bar{\bm{b}}$ incorporating length input is 512 and
        the number of possible lengths $L$ in \textit{LenEmb} is 150.

        We first train our models in supervised ML
        using Adam \cite{adam_2014} as optimizer and anneal the learning rate by a factor of 0.5 every four epochs.
        We also apply gradient clip \cite{clip_2013} with a range of [-10, 10], and batch size is set to 64.

        Then we run RL algorithms on previously trained LC models
        with initial learning rate of 0.00001 and reward $r(\bm{y}^s)$ in RL is also set as the sum of R-1, R-2, and R-L scores.
        During the RL, desired length $l_1^s$ to sample the sentences is average distributed in a interval $[20, 70]$.
        We evaluate the model in validation set at each 2000 iterations and select the model according to its cumulative score of R-1, R-2 and R-L.

        Note that in our experiments, the space is not counted into sentence length which is slightly different
        with \cite{kikuchi_2016}.

\subsection{Experiment Results Analysis}

    \subsubsection{Length Control in ML}
        Although the evaluation score is not the unique objective in this research,
        it is of interest that how exactly the score is deprived by LC capacity.
        The results of four LC models in ML are presented in Table \ref{table2},
        and ROUGE scores are collected with desired length of $45$.
        To embody the accuracy level of our LC models,
        we list several existing summarization baselines
        including ABS, ABS+ \cite{rush_2015}, RAS-LSTM, RAS-Elman and Luong-NMT \cite{chopra_2016}.
        After individually comparing two WLI models and two RLI models,
        we find the two proposed models, \textit{LenLInit} and \textit{LenMC},
        slightly corrupt LC capacity while improve the scores obviously.

        In Table \ref{table1}, we provide a representative example of the summaries
        generated by LC models,
        and results demonstrate that these models are able to
        output well-formed sentences with various lengths.
        It is also observed that \textit{LenLInit} and \textit{LenMC} perform better on short sentence summary in this case.


    \subsubsection{RL for Length Control}
        Table \ref{table3} displays overall comparison of all models under RL.
        We evaluate our models with sentence length of 25, 45 and 65,
        which represent short, median and long sentences separately.
        Results may vary after each training process since RL is usually unstable,
        so we repeat training for multiple times in each model and statistic the results on average.

        We first present the results of four LC models in ML.
        After that, we apply raw self-critical sequence training (SCST) on this basis,
        without any constraints on output length,
        we find that WLI models tend to lose control of length sharply but increase the accuracy significantly,
        while RLI methods still keep the good LC ability.
        This is mainly because the lengths of sampled sentences for RLI models are consistent with the input length in most cases,
        consequently, the training process is stable.

\begin{figure}[t]
    \centering 
    \begin{tabular}{cc}
        \includegraphics[width=0.45\columnwidth]{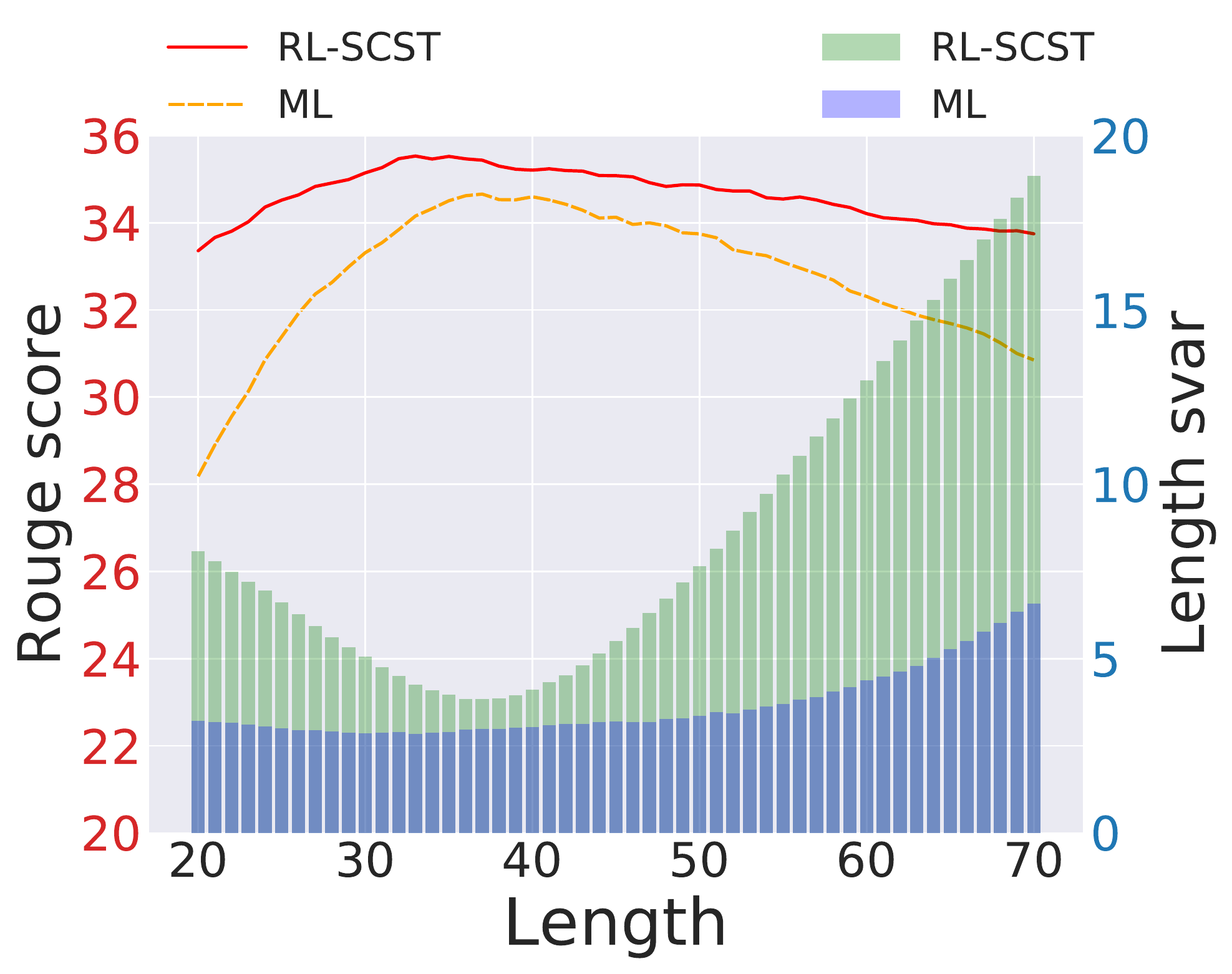} & \includegraphics[width=0.45\columnwidth]{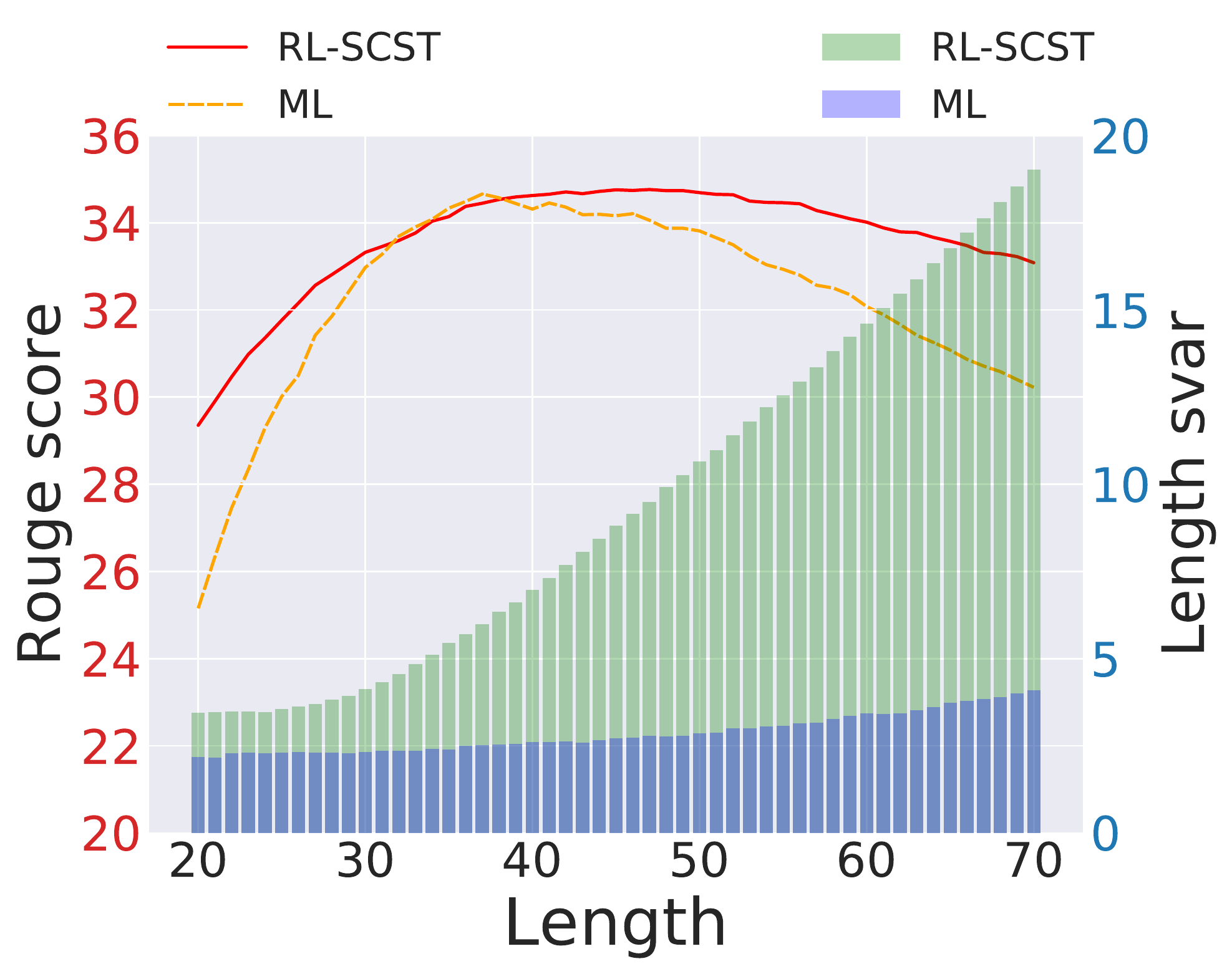} \\
        (a) LenLInit & (b) LenInit \\
        \includegraphics[width=0.45\columnwidth]{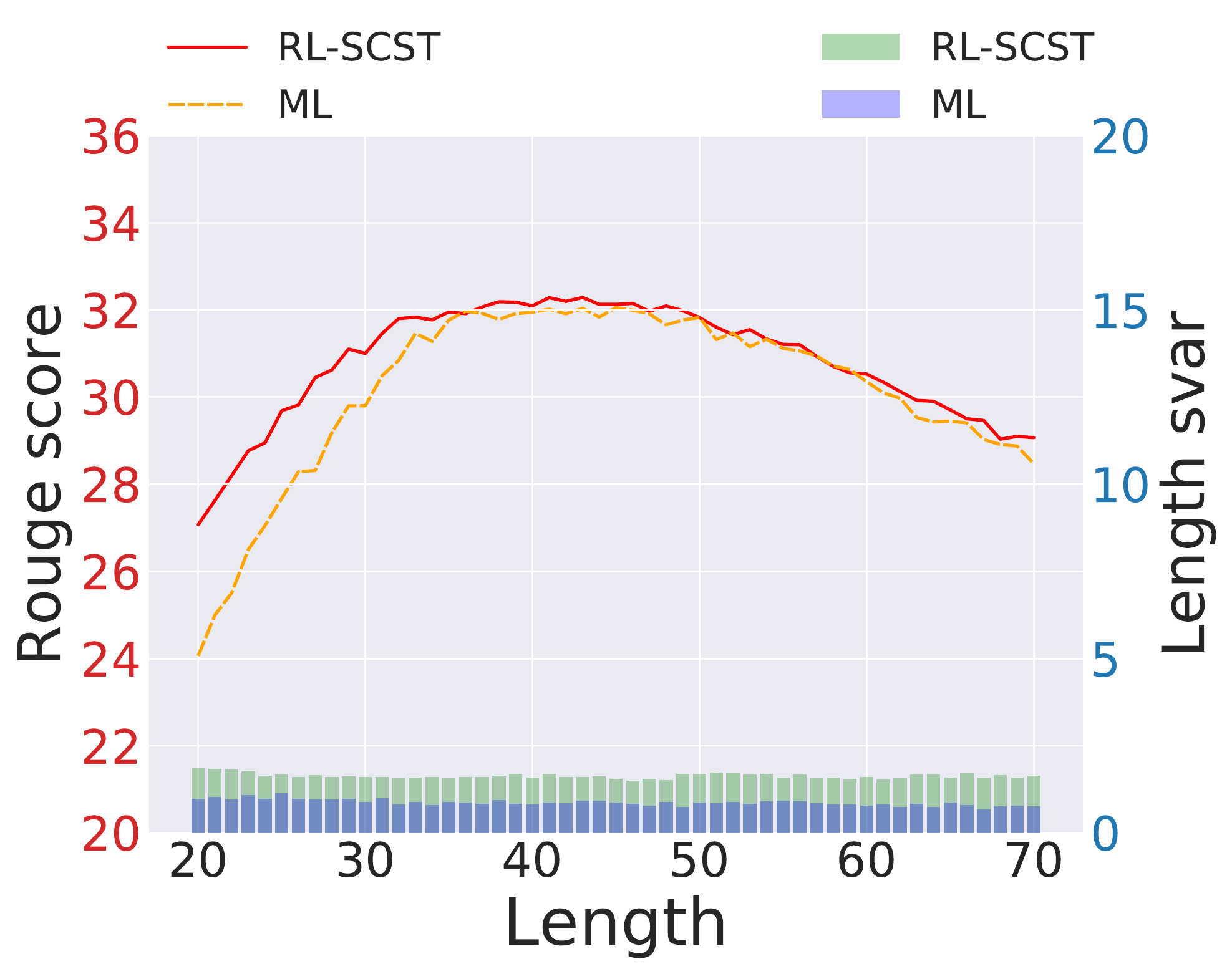} & \includegraphics[width=0.45\columnwidth]{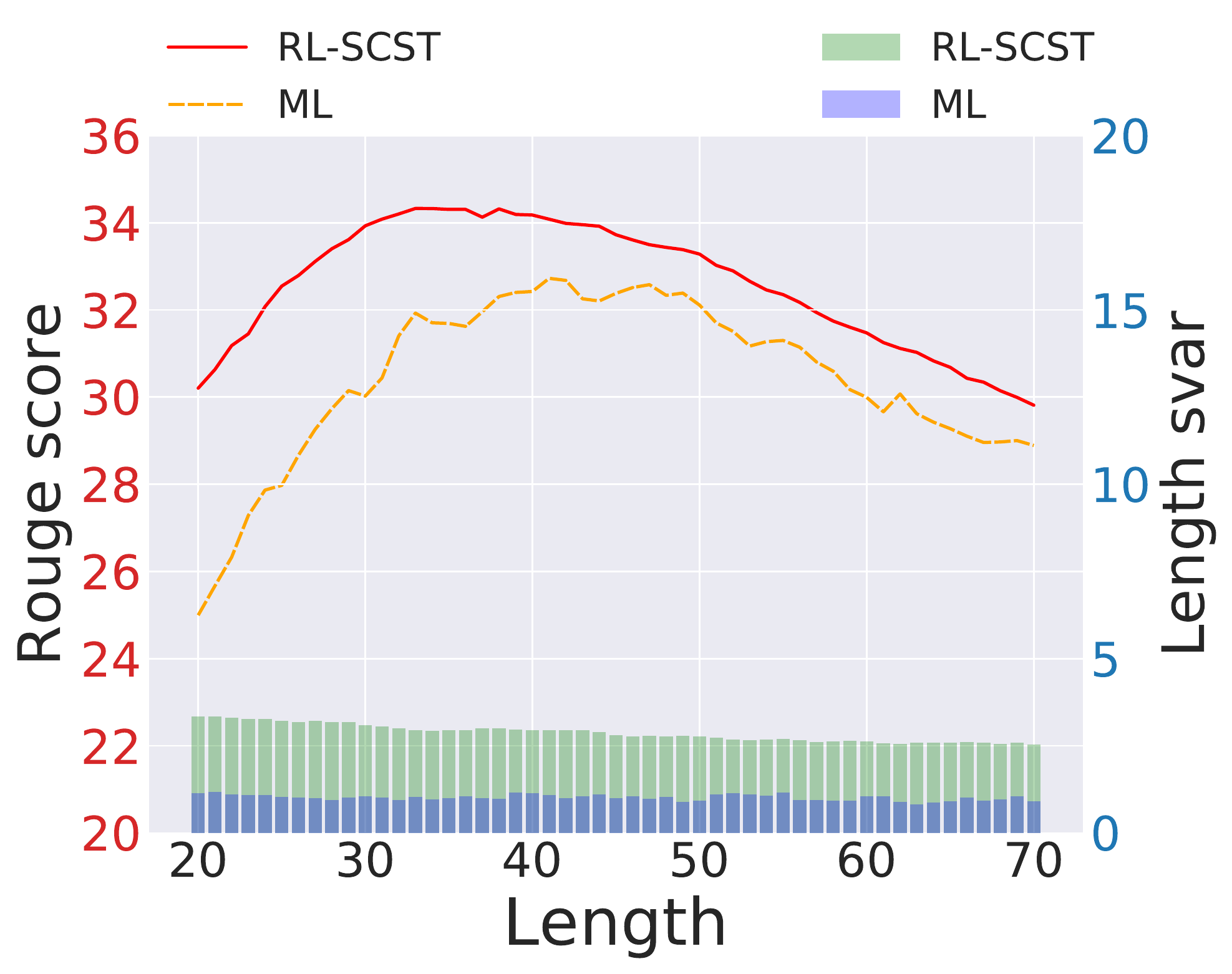} \\
        (c) LenEmb & (d) LenMC
    \end{tabular}
    \caption{
    Performance of SCST-trained LC models versus ML-trained LC models on both aspects of LC capacity (bars) and outputs score (lines).
    }
    \label{fig3}
\end{figure}

        To further investigate the impacts that RL makes on LC models,
        we evaluate the models on all expected lengths within the range of [20, 70].
        These results are reported in Figure \ref{fig3},
        where x-axis represents the length,
        ROUGE score and svar of y-axis measures output quality and LC capacity separately.
        For convenience, we take the average of R-1, R-2, and R-L values as ROUGE score.
        Obviously, RL improves scores among the range of all lengths but release LC capacity.
        The gain of scores is significantly on both short and long sentences for WLI models
        as well as short sentences for RLI models,
        which signify that RL alleviates the problem due to unbalanced amount of multiple lengths in training corpus.
        In particular, \textit{LenLIint} performs the highest score among four models,
        nonetheless, have poor LC on long sentences.
        It is worth noting that \textit{LenMC} with SCST results even higher score than \textit{LenInit}
        on short summaries, and still perserve excellent LC ability.
        Since SCST has negligible effect on \textit{LenEmb},
        we exclude \textit{LenEmb} for further comparison under length-control RL.

    \subsubsection{Controllable Length Control Analysis}

        Results of MTS-RL and SCD-RL in Table \ref{table3} are followed by SCST part.
        We make experiments of MTS-RL on three LC models,
        for WLI models \textit{LenLInit} and \textit{LenInit}, accuracy and svar both rise under the selected $d_{th}$ increasing,
        which means hyper-parameter $d_{th}$ in MTS-RL can be used to adjust the LC capacity.
        However, for the RLI model \textit{LenMC},
        results show there is no obvious distinction in scores when we use different $d_{th}$.
        Hence, we adopt SCD-RL training algorithm for \textit{LenMC},
        the results show our SCD-RL algorithm can control LC capacity for RLI model as MTS-RL does for the WLI model.
        and SCD-RL can also manage the LC capacity for WLI models.
        Overall, two RL training algorithms prevent the model from length control collapsing,
        and make this capacity controllable via their own hyper-parameters.

\begin{figure}[t]
    \centering
    \includegraphics[width=0.9\columnwidth]{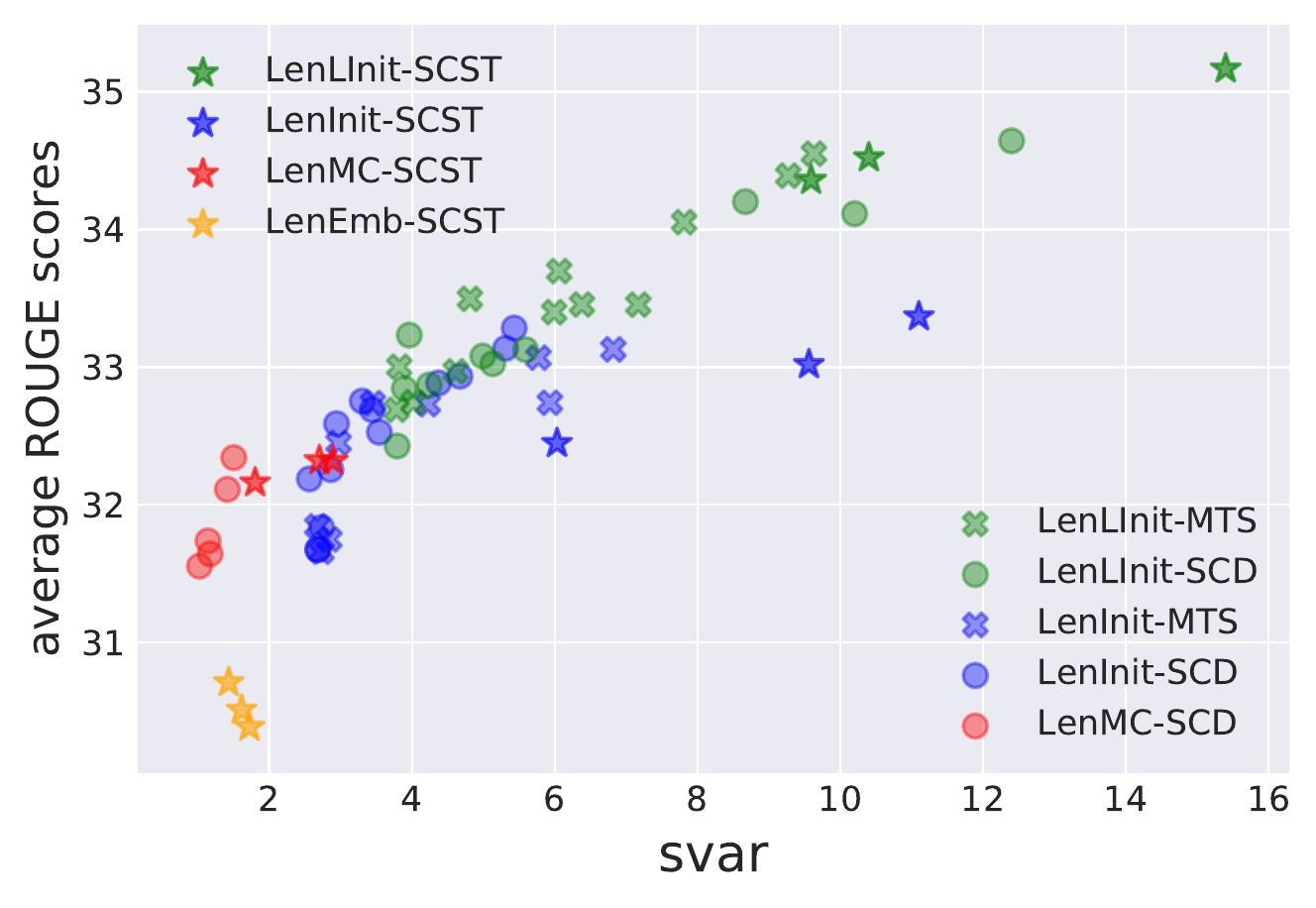} 

    \caption{Exhibition of overall experiment results on four models in length control RL.
      (
      MTS is only applied on WLI models with $d_{th}$ chosen from [4, 8, 10, 16].
      In SCD training,
      $\lambda$ for WLI models is selected from [0.8, 0.4, 0.2, 0.1, 0.05],
      and we set $\lambda$ as one of [0.8, 0.4, 0.1] for \textit{LenMC}.
      )
    }
    \label{fig4}
\end{figure}

        In order to make comprehensive comparison considering all factors,
        we build a scatter map (see Figure \ref{fig4}) to display the performance of models in different training strategies.
        The x-axis is svar to measure the LC capacity.
        To evaluate the scores intergrating different lengths,
        we take the average of R-1, R-2, R-L scores with lengths of
        $25$, $45$, $65$ as the value on y-axis.
        From Figure \ref{fig4}, we can give some intuitive interpretations:
        (i) SCST as length control RL for WLI models is extremely unstable.
        (ii) For those models with similar average ROUGE scores, \textit{LenMC} have strictly better LC capability than \textit{LenInit}.
        (iii) Statistically, \textit{LenLInit} performs higher score than \textit{LenInit} when
        their svar values are relatively close.
        (iv) The models with designed RL algorithms sufficiently cover wide range of LC capacity
        with accuracy in a reasonable scope.


\section{Conclusion and Future Work}
%

In this paper,
we proposed \textit{LenLInit} and \textit{LenMC} inspired by former work,
our modified models improved length control summarization performance on Gigaword Dataset.
Two developed RL algorithms were successfully applied in length control models to significantly improve the scores on all short, median and long sentences,
and to allow users to determine the model with expected length control capacity.
Due to the deficiency of the research in this field, extra work need to be pursued.
We plan to perform experiments on other tasks such as image caption and dialogue system to further verify our RL algorithms.
It is also valuable to investigate the mathematical relationship between length control capacity and evaluation scores,
which can be beneficial for model selection.
Furthermore, the controllable ability can be extended to other domains like sentiment or style.

\bibliography{main.bib}
\bibliographystyle{aaai}
\end{document}